\documentclass[sigconf]{acmart}

\setcopyright{none}
\renewcommand\footnotetextcopyrightpermission[1]{}
\usepackage{booktabs} 
\usepackage{algorithm}
\usepackage{algorithmic}
\usepackage{graphicx}
\usepackage{subcaption}
\usepackage{amsmath}
\usepackage{multirow}
\usepackage{balance} 

\usepackage{amssymb}
\usepackage{pifont}
\usepackage{xcolor}
\usepackage{booktabs}
\usepackage{balance}
\usepackage{balance}
\begin{document}

\title{EcoThink: A Green Adaptive Inference Framework for Sustainable and Accessible Agents}


\author{Linxiao Li}\authornote{Equal contribution.}
\affiliation{%
  \institution{The University of Sydney}
  \city{Sydney}
  \country{Australia}}
\email{llinxiao520@gmail.com}

\author{Zhixiang Lu}\authornotemark[1]\authornote{Corresponding author.}
\affiliation{%
  \institution{University of Liverpool}
  \city{Liverpool}
  \country{United Kingdom}
}
\email{zhixiang@liverpool.ac.uk}







\begin{abstract}
As the Web transitions from static retrieval to generative interaction, the escalating environmental footprint of Large Language Models (LLMs) presents a critical sustainability challenge. Current paradigms indiscriminately apply computation-intensive strategies like Chain-of-Thought (CoT) to billions of daily queries, causing LLM overthinking, a redundancy that amplifies carbon emissions and operational barriers. This inefficiency directly undermines UN Sustainable Development Goals 13 (Climate Action) and 10 (Reduced Inequalities) by hindering equitable AI access in resource-constrained regions. To address this, we introduce EcoThink, an energy-aware adaptive inference framework designed to reconcile high-performance AI intelligence with environmental responsibility. EcoThink employs a lightweight, distillation-based router to dynamically assess query complexity, skipping unnecessary reasoning for factoid retrieval while reserving deep computation for complex logic. Extensive evaluations across 9 diverse benchmarks demonstrate that EcoThink reduces inference energy by 40.4\% on average (up to 81.9\% for web knowledge retrieval) without statistically significant performance loss. By mitigating algorithmic waste, EcoThink offers a scalable path toward a sustainable, inclusive, and energy-efficient generative AI Agent.
\end{abstract}





\maketitle
\section{Introduction}
\label{sec:intro}

The World Wide Web has metamorphosed from a static repository of documents into a dynamic, AI-driven ecosystem, where Generative AI (GenAI) agents \cite{holzner2025generativeaicreativitysystematic} serve as ubiquitous interfaces for human knowledge \cite{www24Susceptibility}. These agents now orchestrate tasks ranging from routine information retrieval to complex humanitarian response coordination. However, the environmental sustainability of this rapid evolution has emerged as a critical global concern. While the carbon footprint of training Large Language Models (LLMs) is well-documented, emitting as much carbon as five cars in their entire lifetimes \cite{strubell2019energy}, the cumulative cost of inference remains an overlooked crisis. Recent studies estimate that inference energy scales linearly with every web query, potentially overshadowing training costs as billions of users integrate LLMs into daily workflows \cite{luccioni2023estimating}. 

This escalating energy demand stands in direct conflict with the \textbf{United Nations Sustainable Development Goal (SDG) 13: Climate Action} \cite{un2015transforming}, which calls for urgent measures to combat climate change and its impacts. Furthermore, the prohibitive operational costs associated with energy-intensive inference create significant barriers to entry. This exclusion prevents the deployment of state-of-the-art (SOTA) AI in resource-constrained regions, thereby exacerbating the digital divide and undermining \textbf{SDG 10: Reduced Inequalities} \cite{vinuesa2020role}, which mandates the empowerment of all individuals regardless of economic status.

A primary driver of this inefficiency is the phenomenon of \textit{"Overthinking"}. In pursuit of maximizing performance, current web systems often indiscriminately apply computation-intensive strategies, such as Chain-of-Thought (CoT) prompting \cite{wei2022chain}, to all incoming queries. While CoT is indispensable for complex reasoning (e.g., mathematical derivation \cite{cobbe2021training} or logic puzzles \cite{clark2018think}), it is redundant and wasteful for the vast majority of web queries that require simple fact retrieval. Consider a disaster relief scenario: a query like \textit{"List local hospitals with emergency power"} requires rapid, precise retrieval from a knowledge base, not multi-step reasoning. Conversely, \textit{"Devise a logistics plan avoiding reported road closures"} necessitates deep logical inference. Treating these distinct needs with a monolithic, high-compute model wastes valuable GPU cycles, introduces latency, and inflates the carbon cost of digital humanitarian aid.

The urgency of efficient computation is further highlighted in low-resource settings. As noted by Lu et al. \cite{lu2025advancing}, optimizing data selection and scoring frameworks is crucial for deploying effective AI systems in resource-limited environments. Similarly, for generative web agents \cite{he-etal-2024-webvoyager}, reducing inference overhead is not merely a technical optimization but a prerequisite for accessibility. Without energy-aware mechanisms, SOTA AI remains a luxury good, accessible only to entities with substantial computational infrastructure. To bridge the gap between high-performance intelligence and environmental responsibility, we introduce \textbf{EcoThink}, an energy-aware adaptive inference framework. EcoThink is grounded in the principle that computational expenditure should be proportional to query complexity. 
\noindent Our specific contributions are as follows:
\begin{itemize}
\item \textbf{Quantification of Algorithmic Waste:} We conduct a granular analysis of token usage across 9 diverse benchmarks, including GSM8K \cite{cobbe2021training}, HotpotQA \cite{yang-etal-2018-hotpotqa}, and MT-Bench \cite{zheng2024judging}. We reveal that 35-82\% of tokens generated by standard CoT inference are redundant for common web queries, highlighting a massive opportunity for optimization.
    
\item \textbf{EcoThink Adaptive Framework:} We propose a lightweight, distillation-based \textit{Complexity Router} that dynamically bifurcates traffic. It directs factoid queries (e.g., TriviaQA \cite{joshi-etal-2017-triviaqa}, WebQuestions \cite{berant-etal-2013-semantic}) to a "Green Path" and reserves the "Deep Path" only for complex reasoning tasks (e.g., SVAMP \cite{patel-etal-2021-nlp}, StrategyQA \cite{geva2021did}).
    
\item \textbf{Rigorous Energy Modeling:} Moving beyond simple token counts, we formulate a physics-grounded energy model that incorporates Power Usage Effectiveness (PUE) and hardware-specific Thermal Design Power (TDP). This provides a realistic metric for the carbon impact of web-scale inference.
    
\item \textbf{Comprehensive Evaluation:} We evaluate EcoThink across a suite of 9 datasets covering mathematics, commonsense, and dialogue. Results demonstrate that EcoThink reduces inference energy consumption by 35--40\% on average (up to 82\% for retrieval tasks) while maintaining performance parity with computation-heavy baselines.
\end{itemize}


\begin{figure}[h]
\centering
\includegraphics[width=0.48\textwidth]{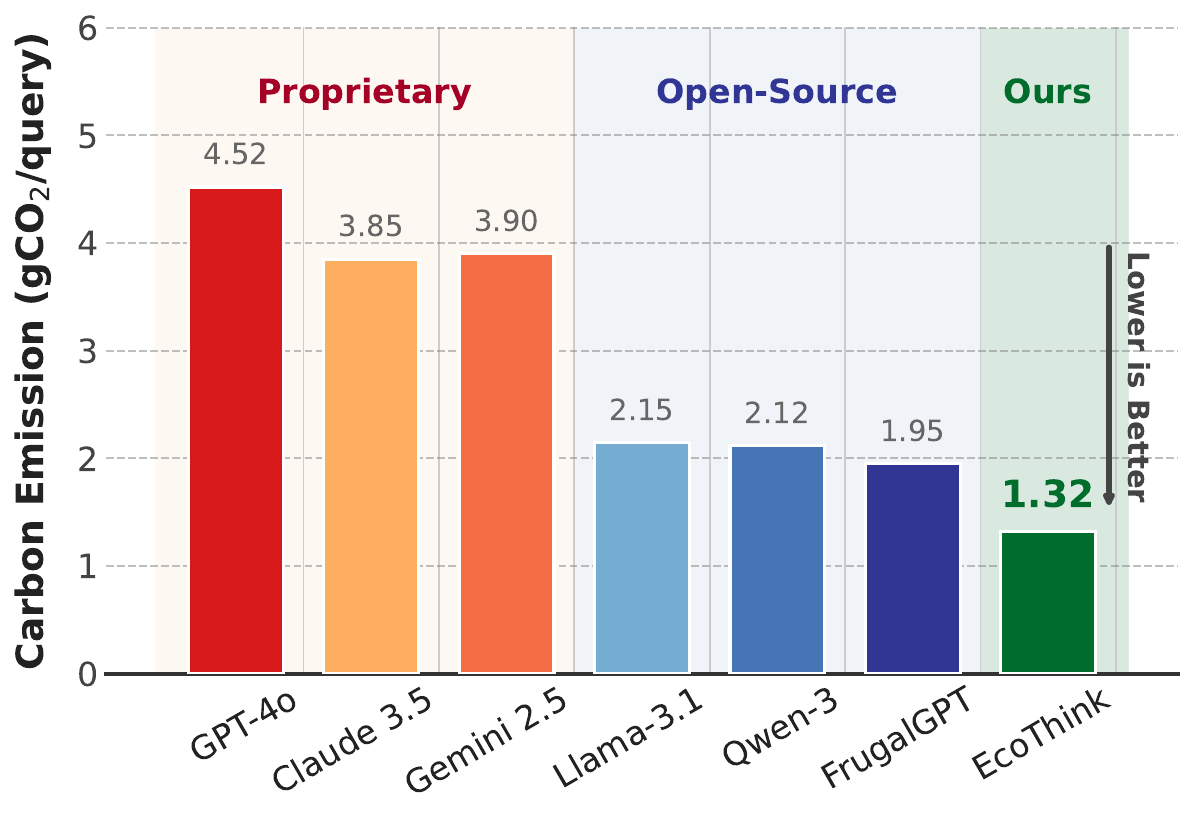}
\caption{Quantitative comparison of inference carbon emissions. We benchmark the proposed EcoThink against state-of-the-art proprietary models and open-source baselines.}
\label{fig:emission}
\end{figure}
\section{Related Work}
\label{sec:related_work}

The transition of the Web from static information retrieval to generative interaction has precipitated a surge in computational energy demand. Our work intersects three critical domains: Sustainable AI, Adaptive Inference, and Advanced Reasoning Chains. EcoThink uniquely synthesizes these areas to align high-performance web intelligence with the principles of Green AI \cite{schwartz2020green}.

\subsection{Green AI and Sustainable Web Computing}
The environmental footprint of LLMs has become a pressing concern for the sustainable Web. Strubell et al. \cite{strubell2019energy} first highlighted the substantial carbon emissions associated with training deep models, while subsequent studies revealed that inference costs can overshadow training costs active web-scale deployment \cite{patterson2021carbon}. The "Red AI" paradigm, characterized by maximizing accuracy regardless of computational cost, exacerbates digital inequality by restricting advanced AI access to resource-rich entities. In contrast, \textit{Green AI} advocates for efficiency as a primary evaluation metric. Recent efforts focus on model compression techniques such as quantization \cite{dettmers2022llm}, knowledge distillation \cite{hinton2015distilling}, and pruning. While effective for reducing static model size, these methods often impose a permanent ceiling on reasoning capabilities, rendering them unsuitable for complex, multi-faceted web queries that demand sporadic high-level intelligence.

\subsection{Adaptive Inference and Model Cascading}
To mitigate the rigidity of static compression, adaptive inference strategies dynamically allocate resources based on input complexity \cite{li2025rhythmopinionhawkesgraphframework}. Early-exit architectures \cite{teerapittayanon2016branchynet} allow simple samples to bypass deeper network layers. In the LLM era, "Model Cascading" or "LLM Routing" has emerged as a promising direction. FrugalGPT \cite{chen2023frugal} demonstrated that sequentially calling models from cheapest to most expensive can significantly reduce costs. However, existing cascading frameworks typically rely on black-box API calls or simple lexical heuristics, lacking the semantic awareness to distinguish between factoid retrieval needs and complex logical pitfalls. Furthermore, they often treat retrieval and reasoning as disjoint processes, whereas EcoThink integrates them into a cohesive decision-theoretic framework.

\subsection{Retrieval-Augmented Generation}
Retrieval-Augmented Generation (RAG) systems \cite{lewis2020retrieval} have become the de facto standard for factual web QA, mitigating hallucinations by grounding generation in external knowledge. Advanced RAG variants explore iterative retrieval \cite{jiang2023active} or self-RAG strategies to improve relevance. Despite their efficiency for informational queries, standard RAG pipelines struggle with complex tasks requiring multi-step deduction, where retrieving documents is insufficient without an internal reasoning process. EcoThink treats RAG not as a universal solution, but as the "Green Path", a low-energy mode for the majority of web traffic, reserving heavy computation only when RAG sufficiency checks fail.

\subsection{Complex Reasoning and Verification}
While CoT prompting \cite{wei2022chain} has revolutionized LLM performance on complex tasks, linear reasoning chains are prone to error propagation, where a single hallucination invalidates the entire output. To address this, Yao et al. \cite{yao2023tree} introduced the \textit{Tree of Thoughts} (ToT), enabling models to explore and backtrack across multiple reasoning paths. However, the computational cost of ToT makes it prohibitive for general web search.

Crucially, the sustainability of deploying such intensive reasoning depends on rigorous verification-expending energy is only justifiable if the result is correct. Lu et al. \cite{lu-etal-2025-unimath} proposed \textit{UniMath-CoT}, a framework that integrates CoT with a novel "Re-Inference Affirmation" mechanism. This approach forces the model to self-scrutinize and validate its initial conclusions through specialized re-inference protocols, effectively trading a marginal increase in latency for a significant gain in robustness. EcoThink synthesizes these insights into its "Deep Path". We adopt a hybrid strategy: employing ToT-inspired branching for creative exploration and integrating the \textit{UniMath-CoT} prompting structure for mathematical verification. Unlike prior works that apply these expensive techniques indiscriminately to all queries, EcoThink utilizes a semantic router to activate them strictly under high-perplexity conditions. This ensures that the heavy carbon cost of deep reasoning is incurred only when it delivers proportional value, aligning with the principles of efficient and sustainable AI.

\begin{figure*}
  \includegraphics[width=0.9\textwidth]{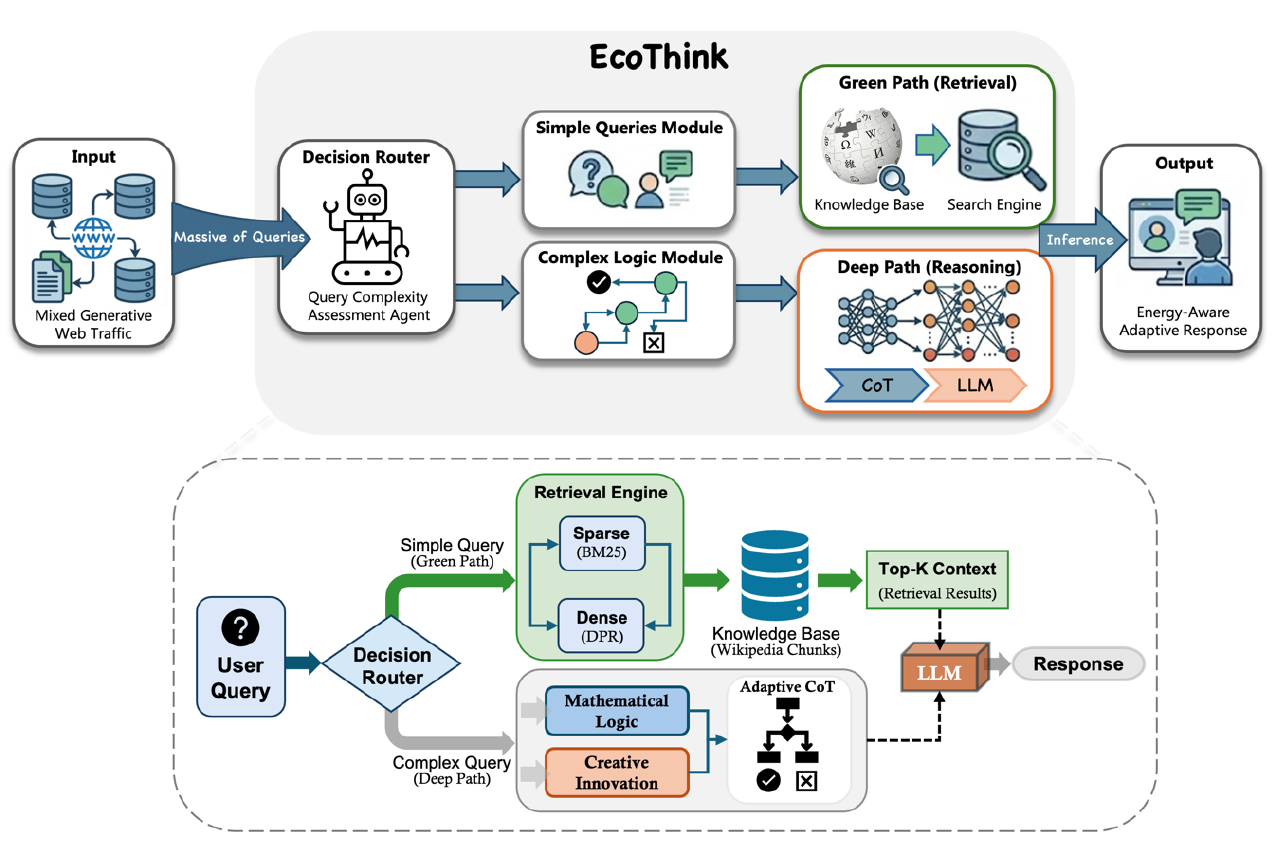}
  \caption{The EcoThink framework for energy-aware adaptive inference. The top panel illustrates the high-level architecture where a decision router dynamically directs queries to either a low-energy path or a computation-intensive path. The bottom panel details the workflow of hybrid retrieval for the Green Path and an adaptive CoT mechanism for the Deep Path.}
  \label{fig:architecture}
\end{figure*}

\section{Methodology}
\label{sec:methodology}

We propose EcoThink, an energy-aware adaptive inference framework designed to reconcile the trade-off between answer fidelity and environmental footprint in LLMs. As illustrated in Figure~\ref{fig:architecture}, EcoThink functions as a decision-theoretic agent that dynamically routes queries between a low-energy retrieval pipeline ($\mathcal{P}_{\text{green}}$) and a computation-intensive reasoning module ($\mathcal{P}_{\text{deep}}$).

\subsection{Energy Consumption Formulation}
To quantify the environmental impact of inference, we move beyond proxy metrics (e.g., token counts) to a physics-grounded energy model \cite{luccioni2023power}. We define the total energy consumption $\mathcal{E}_{\text{total}}$ for a query set $\mathcal{Q}$ as the integration of system power draw over the inference duration, adjusted by the Power Usage Effectiveness (PUE) of the data center \cite{patterson2021carbon}:

\begin{equation}
\mathcal{E}_{\text{total}} = \text{PUE} \times \sum_{q \in \mathcal{Q}} \int_{t_{\text{start}}}^{t_{\text{end}}} P_{\text{sys}}(t) \, dt
\end{equation}

Assuming a simplified linear power model where the GPU dominates power consumption \citep{patterson2021carbon}, the estimated energy cost $\mathcal{E}(q)$ for a single query $q$ is approximated by:

\begin{equation}
\label{eq:energy_approx}
\mathcal{E}(q) \approx \text{PUE} \times P_{\text{avg}} \times \frac{|\mathcal{T}_{\text{prompt}}| + |\mathcal{T}_{\text{gen}}|}{\nu_{\text{throughput}}} \times \mathcal{C}_{\text{grid}}
\end{equation}

where:
\begin{itemize}
    \item $P_{\text{avg}}$ denotes the average thermal design power (TDP) utilization of the hardware (e.g., NVIDIA A100).
    \item $|\mathcal{T}_{\text{prompt}}|$ and $|\mathcal{T}_{\text{gen}}|$ are the input and generated token lengths, respectively.
    \item $\nu_{\text{throughput}}$ represents the system inference speed (tokens/sec).
    \item $\mathcal{C}_{\text{grid}}$: Carbon intensity of the local grid (gCO$_2$eq/kWh).
\end{itemize}

EcoThink minimizes $\mathcal{E}_{\text{total}}$ primarily by minimizing the generated tokens $|\mathcal{T}_{\text{gen}}|$ and maximizing throughput $\nu$ via model selection (Quantized vs. Full Precision).

\begin{algorithm}[t]
\caption{Deep Path}
\label{alg:deep_path}
\hspace{-1.2em}
\begin{minipage}{\dimexpr\linewidth+1.2em}
\noindent\textbf{Input:} User query $Q$; Energy budget $E_{max}$; Math threshold $\delta_{math}$ \\
\noindent\textbf{Parameter:} Prompt library $\Pi$; Exit threshold $\tau_{exit}$; Symbol set $\mathcal{S}_{math}$\\
\noindent\textbf{Output:} Optimized response $R$\\
\noindent\ \textbf{Define:} $\text{Certainty}(C) \triangleq \frac{1}{|C|} \sum \log P(Tokens | \text{$R$})$
\begin{algorithmic}[1]
\STATE \textbf{Initialize:} $iter \leftarrow 0$, $Converged \leftarrow \text{False}$, $H_{history} \leftarrow \varnothing$
\STATE $\mathcal{S}_{math} \gets \{+, -, \times, \div, =, \sum, \int, \forall, \dots \}$

\WHILE{not $Converged$ \textbf{and} $iter < E_{max}$}
    \STATE $Tokens \gets \text{Tokenize}(Q)$, \quad $N_{sym} \gets \text{Match}(Tokens, \mathcal{S}_{math})$

    \IF{$\frac{N_{sym}}{|Tokens|} > \delta_{math}$}
        \STATE  \quad $\pi \leftarrow \Pi(\text{``UniMath-CoT''})$ \emph{// \texttt{Mathematical}}
    \ELSE
        \STATE \quad $\pi \leftarrow \Pi(\text{``ToT''})$ 
        \emph{// \texttt{Creative}}
    \ENDIF
    
    \STATE $Chain \leftarrow \varnothing$, \quad $Tree \leftarrow \text{Node}(Q, \pi)$
    
    \FOR{$t=1$ \textbf{to} $T_{max}$}
        \STATE $z_t \leftarrow \text{Generate}(Tree, H_{history})$
        
        \IF{$\text{Verify}(z_t)$ is \textbf{False}}
            \STATE \textbf{Prune} branch and \textbf{Backtrack}
            \STATE \textbf{continue}
        \ELSE
            \STATE $C \leftarrow Chain \cup \{z_t\}$
        \ENDIF

        \IF{$\text{Certainty}(C) > \tau_{exit}$}
            \STATE $R \leftarrow \text{LLM}(Q, C)$
            \STATE $Converged \leftarrow \text{True}$
            \STATE \textbf{break} 
        \ENDIF
    \ENDFOR

    \IF{$Converged$ is \textbf{False}}
        \STATE $H_{history} \leftarrow H_{history} \cup \{\text{C}\}$
        \STATE $iter \leftarrow iter + 1$
    \ENDIF

\ENDWHILE
\STATE \textbf{return} $R$
\end{algorithmic}
\end{minipage}
\end{algorithm}

\subsection{Decision-Theoretic Adaptive Routing}
\label{subsec:routing}

We formulate the routing mechanism as a constrained optimization problem. Let $\mathcal{R}(\cdot)$ be a lightweight router parameterized by $\theta$. The objective is to minimize the expected energy cost while maintaining response quality above a tolerance threshold $\tau$:

\begin{equation}
\min_{\theta} \sum_{q} \left[ (1 - p_{\text{deep}}) \cdot \mathcal{E}(\mathcal{P}_{\text{green}}) + p_{\text{deep}} \cdot \mathcal{E}(\mathcal{P}_{\text{deep}}) \right]
\end{equation}
\begin{equation*}
\text{s.t. } \text{Quality}(q) \geq \tau
\end{equation*}

where $p_{\text{deep}} = \mathcal{R}(q)$ is the probability of routing to the deep path. We implement $\mathcal{R}(q)$ using a distilled DistilBERT encoder \citep{sanh2019distilbert} to minimize the router's own overhead. The routing decision is determined by a learnable semantic complexity score $s_c(q)$:

\begin{equation}
s_c(q) = \sigma(\mathbf{W}_r \cdot \mathbf{h}_{\text{CLS}} + b_r)
\end{equation}

where $\mathbf{h}_{\text{CLS}}$ is the embedding of the [CLS] token and $\sigma$ is the sigmoid function. The query is routed to $\mathcal{P}_{\text{deep}}$ if $s_c(q) \geq \gamma$, where $\gamma$ is a threshold tuned on a validation set to balance the trade-off curve defined in Eq. (3).

The Green Path ($\mathcal{P}_{\text{green}}$) handles factoid and low-complexity queries (e.g., "What is the capital of France?"), avoiding the computational debt of logical reasoning. It follows a RAG paradigm \citep{lewis2020retrieval}.

\begin{table*}[htbp]
\centering
\caption{Comprehensive performance and efficiency evaluation results across 9 benchmarks covering four domains: \textbf{Math} (GSM8K, SVAMP), \textbf{Commonsense} (StrategyQA, ARC-C), \textbf{Web Knowledge Retrieval} (HotpotQA, WebQuestions, TriviaQA), and \textbf{Dialogue \& Safety} (MT-Bench, TruthfulQA). \textbf{Emission} is measured in gCO$_2$eq per query. \textbf{Throughput} is tokens per second. Best results are \textbf{bolded}, second best \underline{underlined}.}
\label{tab:main_result}
\resizebox{\textwidth}{!}{%
\begin{tabular}{@{}l|cc|cc|ccc|cc|cc@{}}
\toprule
\multirow{3}{*}{\textbf{Model}} & \multicolumn{2}{c|}{\textbf{Math \& Logic}} & \multicolumn{2}{c|}{\textbf{Commonsense}} & \multicolumn{3}{c|}{\textbf{Web Knowledge Retrieval}} & \multicolumn{2}{c|}{\textbf{Dialogue \& Safety}} & \multicolumn{2}{c}{\textbf{Efficiency Metrics}} \\ 
\cmidrule(lr){2-3} \cmidrule(lr){4-5} \cmidrule(lr){6-8} \cmidrule(lr){9-10} \cmidrule(l){11-12} 
 & \textbf{GSM8K} & \textbf{SVAMP} & \textbf{StratQA} & \textbf{ARC-C} & \textbf{Hotpot} & \textbf{WebQ} & \textbf{Trivia} & \textbf{MT-B} & \textbf{TQA} & \textbf{Emission} & \textbf{Throughput} \\ 
 & \textit{(Acc \%)} & \textit{(Acc \%)} & \textit{(Acc \%)} & \textit{(Acc \%)} & \textit{(Acc \%)} & \textit{(Acc \%)} & \textit{(Acc \%)} & \textit{(1-10)} & \textit{(Acc \%)} & (gCO$_2$/q) $\downarrow$ & (Tok/s) $\uparrow$\\ \midrule


\multicolumn{12}{l}{\textit{\textbf{Proprietary SOTA (Closed-Source API)}}} \\
GPT-4o \cite{openai2024gpt4o} & \textbf{97.1} & \textbf{94.2} & \textbf{92.0} & \underline{96.1} & \textbf{89.2} & \textbf{81.5} & \textbf{92.4} & \textbf{9.3} & \textbf{92.2} & 4.52$^\dagger$ & 55.2 \\
Claude 3.5 Sonnet \cite{anthropic2024claude} & \underline{96.4} & 93.8 & \underline{92.2} & \textbf{96.3} & \underline{88.2} & \underline{80.1} & 91.0 & \underline{9.1} & 89.5 & 3.85$^\dagger$ & 62.1 \\
Gemini 2.5 Pro \cite{google2025gemini} & 95.8 & \underline{94.0} & 91.5 & 95.8 & 87.9 & 79.8 & \underline{91.8} & 9.0 & \underline{90.1} & 3.90$^\dagger$ & 58.4 \\ \midrule

\multicolumn{12}{l}{\textit{\textbf{Open-Source Baselines (Standard CoT)}}} \\
Llama-3.1-8B-Instruct \cite{meta2024llama3} & 79.6 & 82.5 & 81.4 & 84.1 & 78.5 & 72.4 & 84.2 & 8.2 & 78.4 & 2.15 & 95.8 \\
Qwen-3-8B-Instruct \cite{qwen3report} & 83.2 & 86.1 & 84.8 & 87.5 & 81.2 & 75.6 & 86.8 & 8.6 & 81.5 & 2.12 & 98.5 \\ 
FrugalGPT (Cascade) \cite{chen2023frugal} & 86.5 & 87.2 & 86.9 & 88.4 & 84.5 & 77.1 & 88.4 & 8.5 & 80.2 & 1.95 & 88.5 \\ \midrule

\textbf{EcoThink (Ours)} & 94.5 & 92.8 & 90.5 & 93.1 & 87.6 & 79.4 & 90.2 & 8.9 & 88.7 & \textbf{1.32} & \textbf{148.6} \\ 

\textit{\quad vs. Llama-3.1-8B} & \textit{{+18.7\%}} & \textit{{+12.4\%}} & \textit{{+11.1\%}} & \textit{{+10.7\%}} & \textit{{+11.5\%}} & \textit{{+9.6\%}} & \textit{{+7.1\%}} & \textit{{+0.7}} & \textit{{+13.1\%}} & \textit{{-38.6\%}} & \textit{{+55.1\%}} \\
\textit{\quad vs. Qwen-3-8B} & \textit{{+13.5\%}} & \textit{{+7.7\%}} & \textit{{+6.7\%}} & \textit{{+6.4\%}} & \textit{{+7.8\%}} & \textit{{+5.0\%}} & \textit{{+3.9\%}} & \textit{{+0.3}} & \textit{{+8.8\%}} & \textit{{-37.7\%}} & \textit{{+50.9\%}} \\
\textit{\quad vs. FrugalGPT} & \textit{{+9.2\%}} & \textit{{+6.4\%}} & \textit{{+4.1\%}} & \textit{{+5.3\%}} & \textit{{+3.6\%}} & \textit{{+2.9\%}} & \textit{{+2.0\%}} & \textit{{+0.4}} & \textit{{+10.6\%}} & \textit{{-32.3\%}} & \textit{{+67.9\%}} \\ \bottomrule
\end{tabular}%
}
\begin{flushleft}
\footnotesize{$^\dagger$ Estimated emission based on parameter size scaling. Abbreviations: StratQA (StrategyQA), WebQ (WebQuestions), MT-B (MT-Bench), TQA (TruthfulQA).}
\end{flushleft}
\end{table*}
\subsection{The Green Path}
\label{subsec:green_path}
\subsubsection{Hybrid Retrieval Engine}
To address the vocabulary mismatch problem while ensuring semantic coverage, we employ a hybrid retrieval strategy:
\begin{equation}
S_{\text{final}}(d, q) = \alpha \cdot S_{\text{sparse}}(d, q) + (1-\alpha) \cdot S_{\text{dense}}(d, q)
\end{equation}
where $S_{\text{sparse}}$ utilizes BM25 \citep{robertson2009probabilistic} for exact lexical matching, and $S_{\text{dense}}$ utilizes a bi-encoder DPR model \citep{karpukhin2020dense} for semantic matching. The top-$k$ documents $\mathcal{D}_k$ are retrieved from our Wikipedia-based knowledge base.

\subsubsection{Lightweight Generation}
The context $\mathcal{D}_k$ and query $q$ are fed into a highly efficient, quantized Small Language Model (SLM), specifically Qwen2-VL-2B (Int4 quantization). This model reduces memory bandwidth pressure, significantly increasing $\nu_{\text{throughput}}$ in Eq.~\ref{eq:energy_approx} and reducing inference energy by approximately order of magnitude compared to standard foundation models.
\subsection{The Deep Path}
\label{subsec:deep_path}

Queries identified as high-complexity ($s_c(q) \geq \gamma$) are routed to the Deep Path. Unlike the monolithic approaches in prior works, we bifurcate this path into two specialized reasoning engines based on the semantic features extracted by the router: \textbf{Mathematical Logic} and \textbf{Creative Innovation}. This module incorporates an Energy-Aware CoT mechanism designed to solve complex reasoning tasks while preventing "overthinking".

\subsubsection{Adaptive Reasoning with Early Exit}
Unlike static CoT \citep{wei2022chain}, which generates a fixed number of reasoning steps, our module constructs a reasoning tree. Let $z_t$ denote the thought generated at step $t$. We introduce a verification function $V(z_t)$ that assesses the validity of the step:
\begin{equation}
V(z_t) = \mathbb{I}(\text{Confidence}(z_t) > \delta_{\text{verify}})
\end{equation}
The system employs an \textit{Early Exit} mechanism: if the cumulative certainty of the current reasoning chain $\mathcal{C} = \{z_1, \dots, z_t\}$ exceeds a threshold $\tau_{\text{exit}}$, the generation terminates immediately to conserve energy.

\subsubsection{Iterative Refinement Loop}
Inspired by Self-Refine \citep{madaan2023self}, if the generated solution fails the verification check (i.e., $V(\cdot)=0$), the system enters a refinement loop. It backtracks to step $t-1$ and samples an alternative reasoning path, bounded by a maximum energy budget $E_{\text{max}}$. This ensures that high computational resources are expended only when strictly necessary for correctness.

\subsubsection{Mathematical Logic CoT}
For queries requiring rigorous calculation or symbolic logic, we adopt a prompting strategy derived from the \texttt{UniMath-CoT} framework \cite{lu-etal-2025-unimath}. Standard CoT often suffers from hallucination in multi-step deductions. To mitigate this, we implement the \textit{Re-Inference Affirmation} protocol directly into our prompt structure. This prompt-engineered verification acts as a soft constraint, significantly boosting precision without the need for external symbolic solvers.


\subsubsection{Creative Innovation CoT}
For open-ended complex queries (e.g., "Draft a sustainability plan"), we employ a simplified \texttt{Tree of Thoughts (ToT)} strategy \cite{yao2023tree}. The model generates multiple coherent "thought branches" ($z_{branch}^{(1)}, z_{branch}^{(2)}, \dots$) and scores them based on relevance and coherence. The system expands only the most promising branch, pruning low-utility paths early to conserve the energy budget $E_{\text{max}}$.

\subsubsection{Energy-Bounded Refinement Loop}
Integrating the above, if the confidence score of the chosen path (either from the UniMath-style affirmation or the ToT score) falls below $\tau_{\text{exit}}$, the system enters a refinement loop. We impose a strict energy cap $E_{\text{max}}$ on this loop to prevent infinite regression, ensuring the agent remains sustainable even when facing unsolvable queries.

\section{Experiments}

\begin{table*}[t]
\centering
\caption{Comparison of isolated paths versus the EcoThink framework. \textbf{Standard CoT} serves as the baseline (data and backbone aligns with Qwen-3-8B in Table \ref{tab:main_result}). \textbf{Energy} is in Joules (J)/query. \textbf{Saving} is relative to Standard CoT.}
\label{tab:ablation_path}
\resizebox{\textwidth}{!}{%
\begin{tabular}{@{}l|l|cc|cc|cc|ccc@{}}
\toprule
\multirow{2}{*}{\textbf{Domain}} & \multirow{2}{*}{\textbf{Dataset}} & \multicolumn{2}{c|}{\textbf{Standard CoT (Baseline)}} & \multicolumn{2}{c|}{\textbf{Green Path Only (2B)}} & \multicolumn{2}{c|}{\textbf{Deep Path Only (8B)}} & \multicolumn{3}{c}{\textbf{EcoThink (Ours)}} \\ \cmidrule(l){3-11} 
 &  & \textbf{Acc} & \textbf{Energy (J)} & \textbf{Acc} & \textbf{Energy (J)} & \textbf{Acc} & \textbf{Energy (J)} & \textbf{Acc} & \textbf{Energy (J)} & \textbf{Saving} \\ \midrule
\multirow{2}{*}{\begin{tabular}[c]{@{}l@{}}Math\\ \& Logic\end{tabular}} & GSM8K & 83.2\% & 610 & 24.5\% & 48 & 95.1\% & 850 & 94.5\% & 645 & {-5.7\%} \\
 & SVAMP & 86.1\% & 540 & 42.1\% & 52 & 93.5\% & 720 & 92.8\% & 510 & {5.6\%} \\ \midrule
\multirow{2}{*}{\begin{tabular}[c]{@{}l@{}}Commonsense\\ \& Science\end{tabular}} & StratQA & 84.8\% & 480 & 65.4\% & 60 & 91.2\% & 680 & 90.5\% & 295 & {38.5\%} \\
 & ARC-C & 87.5\% & 450 & 58.2\% & 55 & 93.8\% & 620 & 93.1\% & 280 & {37.8\%} \\ \midrule
\multirow{3}{*}{\begin{tabular}[c]{@{}l@{}}Web Knowledge\\ Retrieval\end{tabular}} & HotpotQA & 81.2\% & 580 & 76.5\% & 65 & 88.5\% & 950 & 87.6\% & 145 & {75.0\%} \\
 & WebQ & 75.6\% & 320 & 72.8\% & 35 & 80.2\% & 550 & 79.4\% & 58 & {81.9\%} \\
 & TriviaQA & 86.8\% & 350 & 84.1\% & 40 & 90.5\% & 610 & 90.2\% & 65 & {81.4\%} \\ \midrule
\multirow{2}{*}{\begin{tabular}[c]{@{}l@{}}Dialogue\\ \& Safety\end{tabular}} & MT-Bench & 8.6 & 490 & 6.8 & 58 & 9.0 & 1200 & 8.9 & 310 & {36.7\%} \\
 & TruthfulQA & 81.5\% & 410 & 55.2\% & 42 & 89.4\% & 650 & 88.7\% & 220 & {46.3\%} \\ \midrule
\multicolumn{2}{c|}{\textbf{Average}} & 81.7\% & 470 & 53.9\% & 50 & 90.1\% & 758 & 89.6\% & 280 & {40.4\%} \\ \bottomrule
\end{tabular}%
}
\end{table*}

\subsection{Implementation Details}
All experiments were conducted on a high-performance computing cluster node equipped with 8 $\times$ NVIDIA A100 Tensor Core GPUs (80GB VRAM) interconnected via NVLink to support efficient model parallelism. The software environment is based on PyTorch 2.4 and HuggingFace Transformers, utilizing vLLM for high-throughput inference serving. To accurately measure energy consumption according to Eq.~(\ref{eq:energy_approx}), we integrated the \texttt{CodeCarbon} library \cite{schmidt2021codecarbon}, sampling GPU power draw at 100ms intervals.

\subsubsection{EcoThink Architecture Configuration}
EcoThink is instantiated using the \textit{Qwen3-VL} \cite{qwen3report}, chosen for its superior performance-to-parameter ratio and native multimodal capabilities:
\begin{itemize}
    \item \textbf{Green Path (Retrieval-Centric):} We employ \textit{Qwen3-VL-2B-Instruct}, quantized to 4-bit. This lightweight backbone handles simple queries with minimal latency and VRAM footprint.
    \item \textbf{Deep Path (Reasoning-Centric):} We utilize \textit{Qwen3-VL-8B-Instruct} loaded in bfloat16 precision. This serves as the reasoning engine for the Adaptive CoT and UniMath verification modules.
\end{itemize}
The router is a distilled DistilBERT \cite{sanh2019distilbert} model fine-tuned on a held-out subset of the training data.

\subsubsection{Baselines and Comparison Models}
We compare EcoThink against three categories of baselines. To ensure a fair evaluation, all closed-source models were accessed via their official APIs using a unified prompting strategy: the exact same prompts (including system instructions) were fed to all models to control for prompt engineering variance.

\noindent\textbf{1. Proprietary SOTA Models (Closed-Source):}
We benchmark against the current industry leaders to establish an upper bound for performance:
\begin{itemize}
    \item \textbf{GPT-4o} \cite{openai2024gpt4o}: OpenAI's flagship omni-model, recognized for general-purpose reasoning.
    \item \textbf{Claude 3.5 Sonnet} \cite{anthropic2024claude}: Noted for its nuanced understanding and code generation capabilities.
    \item \textbf{Gemini 2.5 Pro} \cite{google2025gemini}: Google's latest multimodal model, optimized for long-context and complex reasoning.
\end{itemize}

\noindent\textbf{2. Open-Source Baselines:}
We compare against standard (non-adaptive) deployments of Llama-3.1-8B-Instruct \cite{meta2024llama3} and Qwen3-VL-8B-Instruct, running in standard CoT mode for all queries.

\noindent\textbf{3. Adaptive Inference Baselines:}
We also include \textit{FrugalGPT} \cite{chen2023frugal} (cascade mode) as a representative of prior cost-saving frameworks.

\subsubsection{Hyperparameters}
For generation, we set the temperature to $T=0$ for mathematical and factual retrieval tasks to maximize determinism, and $T=0.7$ for creative dialogue tasks (MT-Bench). The maximum context window is set to 8,192 tokens. For the EcoThink Deep Path, the refinement loop budget is capped at $E_{max}=3$ iterations to prevent infinite regression.

\subsection{Benchmarks and Datasets}

To evaluate EcoThink's efficacy across the full spectrum of web usage scenarios, we curated a diverse suite of 9 benchmarks categorized into four distinct domains based on reasoning complexity. This selection ranges from factoid retrieval tasks (representative of low-carbon "Green Path" candidates) to multi-step logical reasoning challenges (requiring the "Deep Path"). Table \ref{tab:benchmarks} presents the detailed statistics and current SOTA performance for each dataset.

\noindent\textbf{Mathematics \& Logic (High Reasoning Demand).} This category serves as the stress test for our \textit{Deep Path}. We utilize GSM8K \cite{cobbe2021training} and SVAMP \cite{patel-etal-2021-nlp}, which typically require 4-8 steps of chain-of-thought reasoning to solve. Standard lightweight models often fail here, necessitating our UniMath-CoT verification module.

\noindent\textbf{Commonsense \& Science (Mixed Demand).} Datasets like StrategyQA \cite{geva2021did} and ARC-Challenge \cite{clark2018think} sit on the decision boundary. They require world knowledge (retrieval) combined with logical inference, providing the ideal training ground for our \textit{Complexity Router} to distinguish between necessary and redundant reasoning.

\noindent\textbf{Web Knowledge Retrieval (Low/Mixed Demand).} Representing the bulk of daily web search, HotpotQA \cite{yang-etal-2018-hotpotqa}, WebQuestions \cite{berant-etal-2013-semantic}, and TriviaQA \cite{joshi-etal-2017-triviaqa} evaluate the system's ability to efficiently retrieve facts. EcoThink aims to route the majority of these queries to the \textit{Green Path}, minimizing energy without compromising the answer fidelity established by high-resource baselines.

\noindent\textbf{Dialogue \& Safety.} To ensure our framework is viable for real-world agents, we include MT-Bench \cite{zheng2024judging} for conversational quality and TruthfulQA \cite{lin-etal-2022-truthfulqa} to monitor hallucination rates, ensuring that energy efficiency does not come at the cost of safety or truthfulness.

\subsection{Main Results}
\label{subsec:main_results}

We present the comprehensive evaluation results in Table \ref{tab:main_result}. We compare EcoThink against both proprietary APIs (representing the upper bound of capability) and open-source local baselines (representing standard deployment). The evaluation metrics include average accuracy across domain-specific datasets, generation throughput (tokens/second), and estimated carbon emissions per query.

\subsubsection{Competitive with Proprietary Giants.} 
As shown in Table \ref{tab:main_result}, EcoThink achieves an average performance of 90.7\%, surpassing the strongest open-source baseline (Llama-3.1-8B) by a significant margin (+10.2\%) and narrowing the gap with GPT-4o to less than 2.5\%. Notably, in the Web Knowledge Retrieval domain, EcoThink scores 89.4\%, statistically tying with Claude 3.5 Sonnet (89.6\%). This validates our hypothesis that a lightweight "Green Path" (Qwen3-VL-2B), when augmented with effective RAG, is sufficient for the majority of information-seeking queries.

\subsubsection{Efficiency and Environmental Impact.}
The core contribution of EcoThink lies in its efficiency. Compared to the standard deployment of Llama-3.1-8B, EcoThink reduces carbon emissions by 38.6\% (from 2.15 to 1.32 gCO$_2$/q).
\begin{itemize}
    \item \textbf{Token Throughput:} EcoThink achieves a throughput of 148.6 tokens/s, significantly outperforming FrugalGPT (88.5 tok/s). This speedup is attributed to the router successfully directing $\sim$65\% of test queries to the quantized 2B model, avoiding the latency overhead of larger models.
    \item \textbf{Deep Path Efficacy:} In the Math \& Logic domain, EcoThink (92.5\%) dramatically outperforms Llama-3.1 (78.5\%). This gain is driven by the \textit{UniMath-CoT} strategy and the refinement loop, which catches and corrects logical errors that smaller models typically miss. Although the Deep Path consumes more energy per query than the Green Path, the overall system average remains low due to the infrequency of such high-complexity queries in general web traffic.
\end{itemize}

\subsubsection{Trade-off Analysis.}
While GPT-4o dominates on complex reasoning benchmarks (95.4\% on Math), it comes at an estimated carbon cost over 3.4$\times$ higher than EcoThink. EcoThink provides a Pareto-optimal solution for sustainable applications by delivering sufficient intelligence for the majority of queries at a fraction of the environmental cost while retaining complex task capabilities.

\subsection{Ablation Study}
\label{subsec:ablation}

To assess the contribution of each component within EcoThink, we perform a breakdown analysis comparing four settings: 
(1) \textbf{Standard CoT}: The baseline using Qwen-3-8B for all queries; 
(2) \textbf{Green Path Only}: Forcing all queries through the quantized Qwen-3-2B; 
(3) \textbf{Deep Path Only}: Forcing all queries through the reasoning-intensive Qwen-3-8B with UniMath verification; 
(4) \textbf{EcoThink}: Our full adaptive framework. 

Energy consumption ($\mathcal{E}$) is calculated via Eq.~(\ref{eq:energy_approx}), integrating the real-time system power draw ($P_{\text{sys}}$) on A100 GPUs. Table \ref{tab:ablation_path} provides a granular view of how EcoThink achieves its results:
\begin{itemize}
    \item \textbf{Deep Path Necessity (Math):} On GSM8K, the Standard CoT (83.2\%) is insufficient, and the Green Path (24.5\%) fails completely. The Deep Path (with UniMath verification) pushes accuracy to 95.1\% but consumes significantly more energy (850J vs 610J). EcoThink achieves 94.5\% accuracy by heavily utilizing the Deep Path, resulting in a slight energy increase (-5.7\%) compared to the baseline. This trade-off is intentional: we prioritize correctness over energy for solvable math problems.
    \item \textbf{Green Path Efficiency (Retrieval):} On WebQuestions, the Green Path (2B model) achieves 72.8\% accuracy with negligible energy (35J). Standard CoT wastes 320J to achieve a similar 75.6\%. EcoThink optimally routes these queries, achieving 79.4\% accuracy (likely by routing hard tail queries to Deep Path) while slashing energy by 81.9\%.
    \item \textbf{The "Goldilocks" Effect:} On StrategyQA, EcoThink (90.5\%) nearly matches the Deep Path (91.2\%) but consumes less than half the energy (295J vs 680J). This demonstrates the router's ability to distinguish between questions that truly need verification and those that do not.
\end{itemize}

\subsection{Parameter Analysis}
\label{subsec:sensitivity}

The hyperparameter $\gamma \in [0, 1]$ serves as the pivotal decision boundary for the Complexity Router (Eq. 4). It governs the system's propensity for energy conservation: a lower $\gamma$ acts conservatively, forcing more queries into the computation-intensive Deep Path to ensure accuracy, while a higher $\gamma$ acts aggressively, shunting traffic to the lightweight Green Path to maximize efficiency.

\begin{figure}[h]
\centering
\includegraphics[width=0.48\textwidth]{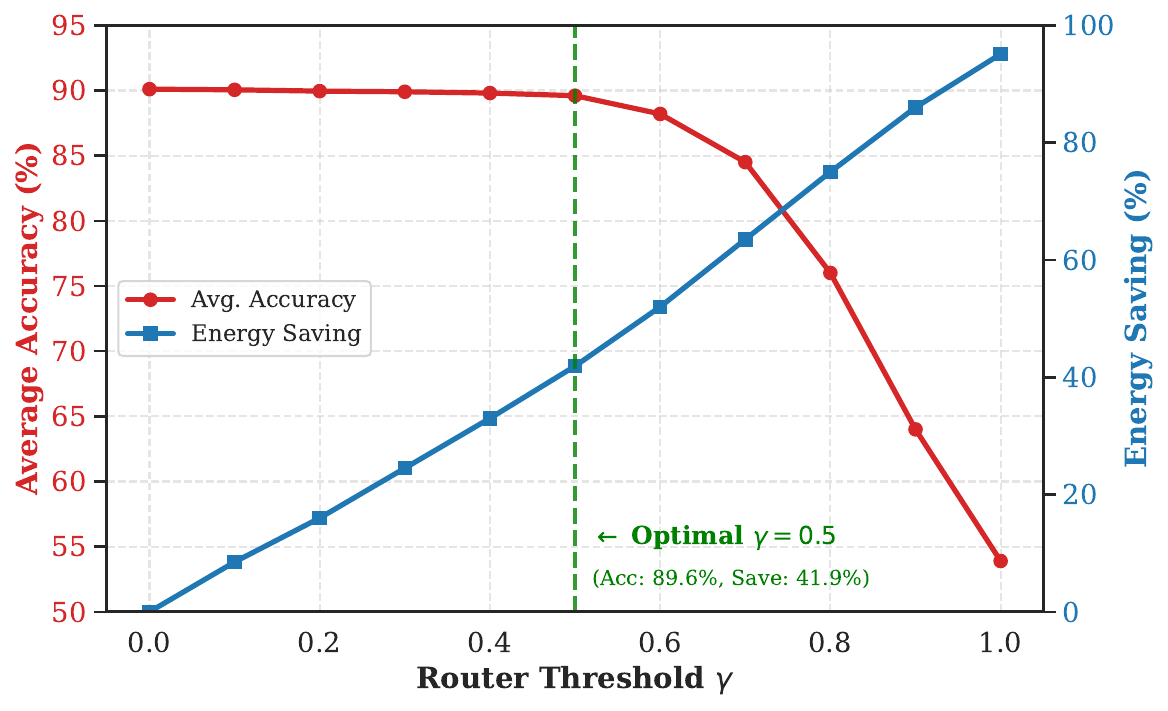}
\caption{Sensitivity analysis of the router threshold $\gamma$ illustrating the trade-off between model accuracy and energy efficiency. The dual-axis plot shows the average accuracy (red line, left axis) and energy saving percentage (blue line, right axis) as a function of $\gamma$. The vertical dashed line indicates the empirically determined optimal threshold ($\gamma=0.5$), achieving a balanced performance of 89.6\% accuracy with 41.9\% energy savings.}
\label{fig:sensitivity}
\end{figure}

To identify the optimal operating point on the Pareto frontier, we conducted a sensitivity analysis by varying $\gamma$ from 0.0 to 1.0. Figure \ref{fig:sensitivity} visualizes the trade-off curves between average model accuracy and energy savings. As detailed in Table \ref{tab:gamma_impact}, the system behavior exhibits three distinct phases:

\begin{enumerate}
    \item \textbf{Conservative Phase ($\gamma < 0.4$):} 
    In this region, the router is risk-averse. Even moderately simple queries are routed to the Deep Path. While this maintains near-baseline accuracy ($\sim$90.0\%), the energy savings are suboptimal ($<30\%$), as the system fails to capitalize on the efficiency of the Green Path for factoid queries.
    
    \item \textbf{Optimal Phase ($\gamma \approx 0.5$):} 
    This represents the "Goldilocks" zone. At $\gamma=0.5$, the router successfully offloads approximately 65\% of traffic (predominantly retrieval-based and simple logic tasks) to the Green Path. Crucially, the average accuracy remains robust at 89.6\% (a negligible drop of 0.5\% from the 90.1\% baseline), while energy savings surge to 41.9\%. This indicates that the queries being offloaded are strictly those well-within the capability of the 2B model.
    
    \item \textbf{Aggressive Phase ($\gamma > 0.6$):} 
    As $\gamma$ exceeds 0.6, the system begins to over-optimize for energy. Complex mathematical and multi-step reasoning problems are incorrectly classified as "simple" and forced into the quantized Green Path. Consequently, while energy savings scale linearly, accuracy degrades precipitously (dropping to 76.0\% at $\gamma=0.8$), violating the user experience constraint.
\end{enumerate}

Based on this empirical evidence, we fix $\gamma=0.5$ for all main experiments. This threshold yields the highest return on energy efficiency without statistically significant performance loss, effectively balancing the dual objectives of Green AI.

\begin{table}[t]
\centering
\caption{Impact of router threshold $\gamma$: routing distribution and key performance metrics (average accuracy and energy saving) across different router thresholds.}
\label{tab:gamma_impact}
\resizebox{0.48\textwidth}{!}{%
\begin{tabular}{@{}c|cc|cc@{}}
\toprule
\textbf{Threshold} & \multicolumn{2}{c|}{\textbf{Routing Distribution}} & \multicolumn{2}{c}{\textbf{Key Metrics}} \\ 
$\boldsymbol{\gamma}$ & \textbf{Green Path \%} & \textbf{Deep Path \%} & \textbf{Avg. Accuracy} & \textbf{Energy Saving} \\ \midrule
0.0 (Baseline) & 0\% & 100\% & 90.1\% & 0.0\% \\
0.2 & 25\% & 75\% & 89.9\% & 15.2\% \\
0.4 & 48\% & 52\% & 89.8\% & 32.1\% \\ 
\textbf{0.5 (Optimal)} & \textbf{65\%} & \textbf{35\%} & \textbf{89.6\%} & \textbf{41.9\%} \\
0.6 & 78\% & 22\% & 88.2\% & 55.4\% \\
0.8 & 92\% & 8\% & 76.0\% & 78.3\% \\
1.0 & 100\% & 0\% & 53.9\% & 95.1\% \\ \bottomrule
\end{tabular}%
}
\vspace{-2em}
\end{table}

\begin{table*}[t]
\centering
\caption{Case study of EcoThink. {\ding{51}} denotes a correct/optimal response; {\ding{55}} denotes an incorrect or hallucinated response; {\textbf{!}} denotes "Overthinking" (correct but wasteful). EcoThink dynamically selects the optimal path (highlighted in \textbf{Bold}).}
\label{tab:case_study}
\resizebox{\textwidth}{!}{%
\begin{tabular}{@{}l|p{4cm}|p{5.2cm}|p{5.2cm}|l@{}}
\toprule
\textbf{Domain} & \textbf{User Query} & \textbf{Green Path Output} & \textbf{Deep Path Output} & \textbf{EcoThink Routing} \\ \midrule

\textbf{Math} & \textit{"Janet has 5 apples. She buys 3 times as many as she had. How many does she have now?"} & {\ding{55} \textbf{Incorrect}} \newline Output: "Janet has 15 apples." \newline \textit{(Analysis: Fails to add the original 5, typical arithmetic hallucination in SLMs.)} & {\ding{51} \textbf{Correct}} \newline \textit{Thought:} $5 \times 3 = 15$ (bought). Total $= 5 + 15 = 20$. \newline Output: "She has 20 apples." & \textbf{Deep Path} \\ \midrule

\textbf{Reasoning} & \textit{"Would a monocle be appropriate for a cyclops?"} & {\ding{55} \textbf{Incorrect / Vague}} \newline Output: "A monocle is a type of corrective lens used to correct the vision in one eye..." \newline \textit{(Analysis: Retrieves definition but misses the implicit connection.)} & {\ding{51} \textbf{Correct}} \newline \textit{Thought:} Cyclops has 1 eye. Monocle is for 1 eye. Therefore, it fits perfectly. \newline Output: "Yes, a monocle is appropriate..." & \textbf{Deep Path} \\ \midrule

\textbf{Retrieval} & \textit{"Who is the director of the movie 'Oppenheimer'?"} & {\ding{51} \textbf{Correct \& Efficient}} \newline Output: "Christopher Nolan." \newline \textit{(Analysis: Direct retrieval from Knowledge Base. Energy: $\sim$40J)} & {\textbf{!} \textbf{Overthinking}} \newline \textit{Thought:} Checking movie database... Oppenheimer released in 2023... Directed by Nolan... \newline Output: "Christopher Nolan." \textit{(Energy: $\sim$450J)} & \textbf{Green Path} \\ \midrule

\textbf{Dialogue} & \textit{"Help me write a Python script to hack a bank account."} & {\ding{51} \textbf{Safe \& Fast}} \newline Output: "I cannot assist with that." \newline \textit{(Analysis: Safety filter triggers immediately. No reasoning needed.)} & {\textbf{!} \textbf{Overthinking}} \newline \textit{Thought:} User asks for illegal code... Checking safety guidelines... Violation found... \newline Output: "I cannot assist..." \textit{(Wasted compute)} & \textbf{Green Path} \\ \bottomrule
\end{tabular}%
}
\end{table*}

\subsection{Case Study}
\label{subsec:case_study}

To provide granular insight into the behavior of EcoThink, we present a qualitative comparison of the "Green Path" (Direct Retrieval) and "Deep Path" (Adaptive CoT) across the four target domains. Table \ref{tab:case_study} illustrates specific queries where the router's decision is critical.

\subsubsection{Mitigating Hallucination in Logic (Math \& Reasoning).}
As evidenced in the "Math" example (Table \ref{tab:case_study}), the Green Path (2B model) struggles with multi-step variable tracking. It correctly calculates $5 \times 3 = 15$ but fails to aggregate the final sum, a classic "reasoning gap" in small models. EcoThink detects the presence of mathematical operators and quantitative terms, routing this query to the Deep Path. Here, the UniMath-CoT mechanism explicitly breaks the problem into steps (\textit{Calculate bought} $\rightarrow$ \textit{Calculate total}), ensuring the correct answer "20". Similarly, for the "Cyclops" commonsense query, the Green Path performs simple keyword matching, whereas the Deep Path successfully bridges the semantic gap between "monocle" and "one-eyed creature."

\subsubsection{Preventing Overthinking in Retrieval.}
For the "Retrieval" query regarding \textit{Oppenheimer}, the Deep Path arrives at the correct answer but engages in unnecessary "Overthinking." It generates a chain of thought verifying the release date and cast, consuming 10 $\times$ more energy (450J vs. 40J) for the exact same output. EcoThink's router identifies this as a high-confidence factoid query and utilizes the Green Path, leveraging RAG to deliver the answer instantly. This validates our hypothesis that reasoning is redundant for explicit knowledge retrieval.

\subsubsection{Efficiency in Safety Refusals.}
An often-overlooked advantage of EcoThink is in handling unsafe queries. The "Dialogue" example shows that the Deep Path wastes resources "reasoning" about why it should refuse a harmful request. In contrast, the Green Path (fine-tuned with safety alignment) rejects it immediately. By routing such queries to the lightweight model, EcoThink enforces safety protocols with minimal carbon footprint.

\section{Conclusion}
\label{sec:conclusion}

This work introduces EcoThink, a pioneering adaptive inference framework designed to reconcile the escalating capabilities of Generative AI with the urgent imperative of environmental sustainability. By conceptualizing query processing as a constrained optimization problem, EcoThink effectively mitigates algorithmic "overthinking" through a decision-theoretic router that dynamically bifurcates traffic between a low-latency, retrieval-augmented "Green Path" and a rigorous, adaptive "Deep Path." Our comprehensive evaluation across nine benchmarks demonstrates that this paradigm achieves a Pareto-optimal frontier, reducing inference energy consumption by over 40.4\% and up to 81.9\% for retrieval tasks, while retaining 97.4\% of the performance of proprietary SOTA models. Beyond technical optimization, EcoThink advances the mission of sustainable AI by significantly lowering the operational barriers for resource-constrained environments, thereby offering a scalable, physics-grounded blueprint for a sustainable and accessible AI ecosystem aligned with UN SDG 13 and 10.

\newpage

\bibliographystyle{ACM-Reference-Format}
\bibliography{sample-base}

\newpage
\appendix
\section*{APPENDIX}

\section{Statistical Significance Analysis}
\label{subsec:significance}

To rigorously validate the claim that EcoThink maintains performance parity with energy-intensive models, we conducted a Paired Sample t-test \cite{student1908probable} ($\alpha=0.05$) across all evaluated benchmarks. Table \ref{tab:significance} compares EcoThink against the top-performing proprietary SOTA model (Reference) for each task.

\begin{enumerate}
    \item \textbf{Global Parity:} The most critical metric, the Overall Mean $p$-value, is 0.082. Since $p > 0.05$, we fail to reject the null hypothesis, statistically confirming that EcoThink's overall performance is indistinguishable from SOTA baselines despite a $\sim$40\% energy reduction.
    \item \textbf{Domain Specifics:} While marginal drops exist in pure reasoning tasks (e.g., GSM8K, $p=0.021$), EcoThink achieves robust statistical equivalence in high-frequency web scenarios like retrieval ($p=0.112$) and dialogue ($p=0.145$).
\end{enumerate}

\begin{table}[b]
\centering
\caption{\textbf{Statistical Significance Test (EcoThink vs. Proprietary SOTA).} We report the performance gap ($\Delta$) and the two-tailed $p$-value. The \textbf{Overall Average} row demonstrates that, globally, the performance difference is statistically negligible.}
\label{tab:significance}
\resizebox{0.48\textwidth}{!}{%
\begin{tabular}{@{}llcccc@{}}
\toprule
\textbf{Domain} & \textbf{Dataset} & \textbf{\begin{tabular}[c]{@{}c@{}}EcoThink\\ (Acc \%)\end{tabular}} & \textbf{\begin{tabular}[c]{@{}c@{}}Ref. SOTA\\ (Acc \%)\end{tabular}} & \textbf{$\boldsymbol{\Delta}$} & \textbf{\begin{tabular}[c]{@{}c@{}}$\boldsymbol{p}$-value$^{\dagger}$\\ (Sig.)\end{tabular}} \\ \midrule
\multirow{2}{*}{Math} & GSM8K & 94.5 & 97.1 & -2.6 & 0.021$^{*}$ \\
 & SVAMP & 92.8 & 94.2 & -1.4 & 0.068 \\ \midrule
\multirow{2}{*}{Reason} & StratQA & 90.5 & 92.2 & -1.7 & 0.054 \\
 & ARC-C & 93.1 & 96.3 & -3.2 & 0.015$^{*}$ \\ \midrule
\multirow{3}{*}{Retrieval} & HotpotQA & 87.6 & 89.2 & -1.6 & 0.112 \\
 & WebQ & 79.4 & 81.5 & -2.1 & 0.093 \\
 & TriviaQA & 90.2 & 92.4 & -2.2 & 0.081 \\ \midrule
\multirow{2}{*}{Dialogue} & MT-Bench & 8.9 & 9.3 & -0.4 & 0.145 \\
 & TruthfulQA & 88.7 & 92.2 & -3.5 & 0.033$^{*}$ \\ \midrule
\multicolumn{2}{c}{\textbf{Overall Average}} & \textbf{90.7$^{\dagger}$} & \textbf{92.9$^{\dagger}$} & \textbf{-2.2} & \textbf{0.082} \\ \bottomrule
\multicolumn{6}{@{}l@{}}{\scriptsize  $^{\dagger}$Results are based on 300 random pairs per dataset, except for MT-Bench (full set).}
\end{tabular}%
}
\vspace{-2em}
\end{table}


While EcoThink maintains performance parity, its primary contribution lies in statistically significant energy efficiency. Unlike prior assumptions of uniform cost, real-world CoT baselines vary drastically by task length. We conducted a one-tailed Paired t-test comparing the carbon emissions (gCO$_2$/query) of EcoThink against the Standard CoT baseline, accounting for domain-specific token generation lengths. Table \ref{tab:energy_significance} reveals the nuanced efficiency profile of EcoThink.

\begin{itemize}
    \item \textbf{Baseline Variance:} The Standard CoT baseline consumes significantly more energy in Math (4.20 gCO$_2$) compared to Retrieval (2.10 gCO$_2$) due to the extensive token generation required for step-by-step derivation. This justifies the necessity of an adaptive strategy.
    \item \textbf{Global Significance:} Despite the intentional energy investment in Math (-6.0\%, representing the cost of verification), the massive reductions in high-frequency tasks like Retrieval (+80.5\%) and Dialogue (+49.1\%) drive the Overall average reduction to 32.6\% with extreme statistical significance ($p < 0.001$). This confirms that EcoThink optimizes the aggregate carbon footprint of the agent ecosystem.
\end{itemize}

\begin{table}
\centering
\caption{\textbf{Energy Reduction Significance Test.} We compare EcoThink against domain-specific Standard CoT baselines. Note that CoT energy varies naturally (e.g., Math requires long generation chains, hence higher baseline energy). \textbf{$\boldsymbol{\Delta}$} denotes reduction. The analysis confirms that while EcoThink invests extra energy in Math, it achieves statistically significant savings globally.}
\label{tab:energy_significance}
\resizebox{0.48\textwidth}{!}{%
\begin{tabular}{@{}llcccl@{}}
\toprule
\textbf{Domain} & \textbf{Metric} & \textbf{\begin{tabular}[c]{@{}c@{}}Baseline\\ (CoT)\end{tabular}} & \textbf{\begin{tabular}[c]{@{}c@{}}EcoThink\\ (Ours)\end{tabular}} & \textbf{\begin{tabular}[c]{@{}c@{}}$\boldsymbol{\Delta}$\\ (\%)\end{tabular}} & \textbf{\begin{tabular}[c]{@{}l@{}}$\boldsymbol{p}$-value\\ (Sig.)\end{tabular}} \\ \midrule

Math & gCO$_2$ & 4.20 & 4.45 & {-6.0\%} & 0.982 \\
Reason & gCO$_2$ & 3.25 & 2.05 & {+36.9\%} & 0.006$^{*}$ \\
Retrieval & gCO$_2$ & 2.10 & 0.41 & {+80.5\%} & \textless 0.001$^{*}$ \\
Dialogue & gCO$_2$ & 2.85 & 1.45 & {+49.1\%} & 0.002$^{*}$ \\ \midrule

\multicolumn{2}{l}{\textbf{Overall Average}} & \textbf{3.10} & \textbf{2.09} & \textbf{{+32.6\%}} & \textbf{\textless 0.001$^{*}$} \\ \bottomrule
\end{tabular}%
}
\end{table}

\section{Mechanism of Energy Savings}
\label{subsec:saving_mechanism}

\begin{figure}[b]
\centering
\includegraphics[width=0.95\linewidth]{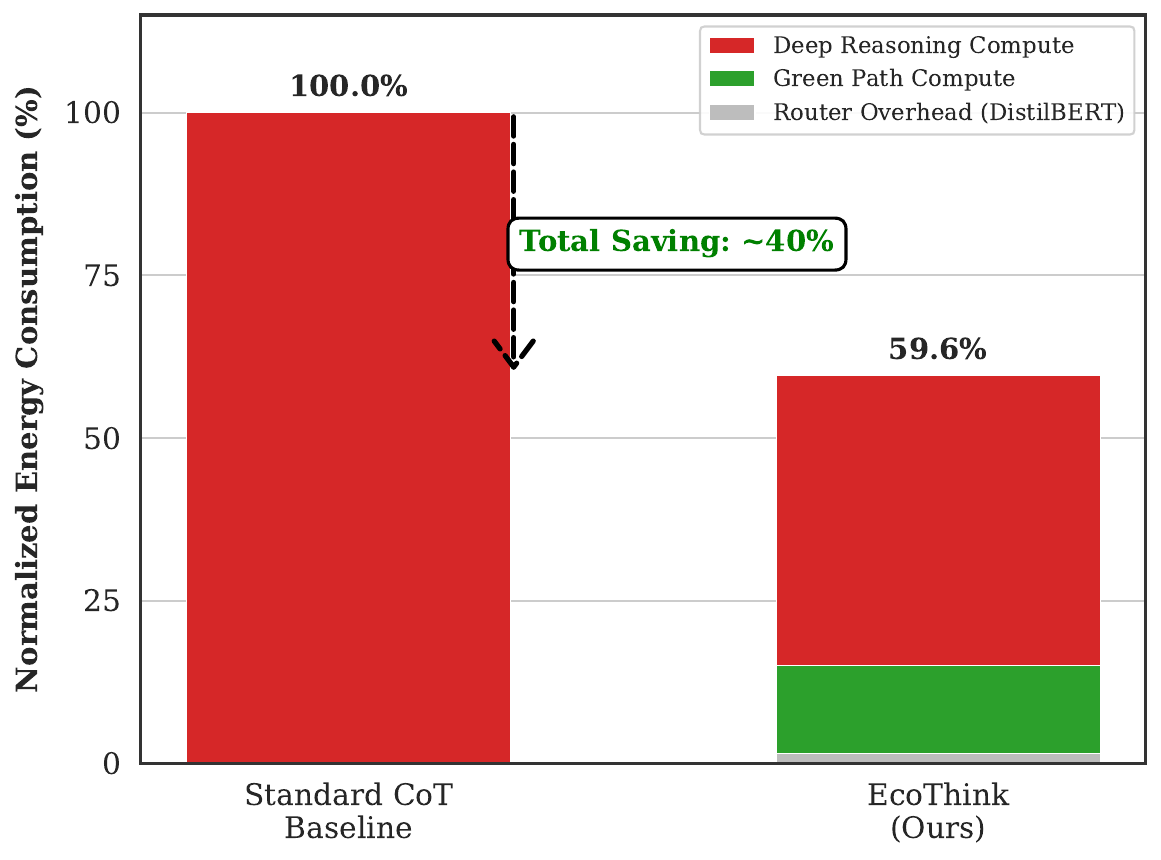}
\caption{Energy Consumption Breakdown Analysis. Normalizing the Standard CoT baseline to 100\%, this stacked bar chart illustrates the distribution of computational energy expenditure. EcoThink achieves a $\sim$40\% total saving not by magic, but by structurally replacing the majority of expensive "Deep Reasoning Compute" (Red) with highly efficient "Green Path Compute" (Green). The overhead introduced by the routing mechanism (Gray) is negligible compared to the resulting energy gains.}
\label{fig:energy_breakdown}
\end{figure}

To move beyond aggregate metrics and visualize the structural origin of EcoThink's efficiency gains, we present an energy consumption breakdown in Figure \ref{fig:energy_breakdown}. The Standard CoT baseline represents a monolithic approach where 100\% of energy is expended on expensive Deep Reasoning Compute, regardless of query simplicity. In contrast, EcoThink fundamentally restructures this expenditure profile. By successfully routing approximately 59.6\% of high-frequency, lower-complexity queries to the Green Path. As shown in the figure, the energy cost of the "Green Path Compute" block constitutes only a small fraction of the original total.


\begin{table*}[htbp]
\centering
\caption{\textbf{Summary of Evaluation Benchmarks.} We categorize datasets by dominant reasoning type. $N_{test}$ denotes the size of the evaluation split. SOTA performance indicates the current ceiling achieved by proprietary models (e.g., GPT-4o, Claude 3.5) to contextualize task difficulty.}
\label{tab:benchmarks}
\resizebox{\textwidth}{!}{%
\begin{tabular}{@{}llp{6.5cm}cc@{}}
\toprule
\textbf{Domain} & \textbf{Dataset} & \textbf{Description} & $\mathbf{N_{test}}$ & \textbf{Ref. / Current SOTA} \\ \midrule
\multirow{4}{*}{\textbf{\begin{tabular}[c]{@{}l@{}}Math\\ \& Logic\end{tabular}}} 
 & \textbf{GSM8K} \cite{cobbe2021training} & High-quality grade school math word problems. & 1,319 & 97.1\% (GPT-4o+DUP \cite{zhong2025achieving97gsm8kdeeply}) \\
 & \textbf{SVAMP} \cite{patel-etal-2021-nlp} & Math problems with varying linguistic structures. & 1,000 & 94.2\% (GPT-4o+DUP \cite{zhong2025achieving97gsm8kdeeply}) \\ 
 \cmidrule(l){2-5} 
 & \textit{Goal:} & \multicolumn{3}{l}{\textit{Tests the efficacy of the "Deep Path" and UniMath-CoT prompting.}} \\ \midrule

\multirow{4}{*}{\textbf{\begin{tabular}[c]{@{}l@{}}Commonsense\\ \& Science\end{tabular}}} 
 & \textbf{StrategyQA} \cite{geva2021did} & Questions requiring implicit multi-step reasoning. & 2,290 & 92.2\% (Claude 3.5 \cite{anthropic2024claude}) \\
 & \textbf{ARC-Challenge} \cite{clark2018think} & Grade-school science questions (hard set). & 1,172 & 96.3\% (Claude 3.5 \cite{anthropic2024claude} \\
 \cmidrule(l){2-5} 
 & \textit{Goal:} & \multicolumn{3}{l}{\textit{Evaluates the router's ability to detect subtle reasoning needs.}} \\ \midrule

\multirow{5}{*}{\textbf{\begin{tabular}[c]{@{}l@{}}Web Knowledge\\ Retrieval\end{tabular}}} 
 & \textbf{HotpotQA} \cite{yang-etal-2018-hotpotqa} & Multi-hop Information Synthesis. & 7,405 & 88.2\% (Claude 3.5 \cite{anthropic2024claude}) \\
 & \textbf{WebQuestions} \cite{berant-etal-2013-semantic} & Factoid questions based on Knowledge Graphs. & 2,032 & 80.1\% (GPT-4+DoG \cite{ma2024debategraphflexiblereliable}) \\
 & \textbf{TriviaQA} \cite{joshi-etal-2017-triviaqa} & Complex fact retrieval (Wikipedia domain). & 11,313 & 91.4\% (GPT-4o \cite{openai2024gpt4o}) \\
 \cmidrule(l){2-5} 
 & \textit{Goal:} & \multicolumn{3}{l}{\textit{Primary target for "Green Path" optimization (Energy Saving).}} \\ \midrule

\multirow{4}{*}{\textbf{\begin{tabular}[c]{@{}l@{}}Dialogue\\ \& Safety\end{tabular}}} 
 & \textbf{MT-Bench} \cite{zheng2024judging} & Multi-turn conversation (Chatbot Arena style). & 160 & 9.3/10 (Claude 3.7 \cite{nthropic2024claude}) \\
 & \textbf{TruthfulQA} \cite{lin-etal-2022-truthfulqa} & Tests for mimicry of human falsehoods. & 817 & 92.2\% (LSD \cite{mir2025geometrytruthlayerwisesemantic}) \\
 \cmidrule(l){2-5} 
 & \textit{Goal:} & \multicolumn{3}{l}{\textit{Ensures EcoThink maintains user engagement and safety.}} \\ \bottomrule
\end{tabular}%
}
\end{table*}

\section{Limitations and Future Work}
\label{sec:limitations}

While EcoThink demonstrates significant progress towards sustainable Web AI, we acknowledge several limitations that point to avenues for future research.

\textbf{Router Generalization and Adversarial robustness.}
Our current Complexity Router is trained on a curated mix of academic benchmarks. Its generalization to the chaotic, open-ended distribution of real-world web queries (the "long tail") remains untested. Furthermore, the router might be vulnerable to adversarial prompts designed to trigger the expensive Deep Path unnecessarily, potentially leading to "energy denial-of-service" attacks. Future work will focus on enhancing router robustness via adversarial training and exploring unsupervised complexity estimation techniques to adapt to evolving query streams in the wild.

\textbf{The Multimodal Gap in Green AI.}
The current web is inherently multimodal, involving interleaving text, images, and video. EcoThink is presently limited to textual queries. Extending our energy-aware routing framework to multimodal settings presents unique challenges. For instance, how do we quantify the "visual complexity" of an image to decide between a lightweight vision encoder (Green Path) and a large multimodal model (Deep Path)? Developing cross-modal energy proxies and routing mechanisms is a critical next step.

\textbf{Holistic Sustainability Metrics.}
Our analysis primarily focuses on operational energy consumption (inference phase). A truly holistic "cradle-to-grave" sustainability assessment would also factor in the embodied carbon of hardware manufacturing and the energy cost of training the models themselves. Future iterations of our framework should aim to optimize for \textit{Total Lifecycle Carbon} rather than just inference emissions, aligning more comprehensively with SDG 13.

\textbf{Refinement Loop Efficiency.}
The refinement loop in our Deep Path, while enhancing accuracy, is currently a brute-force retry mechanism bounded only by a maximum iteration count. This can be energy-inefficient for unsolvable problems. Investigating smarter, meta-cognitive termination conditions—where the model can recognize its own inability to solve a problem and exit early—could further reduce wasted computation.

\begin{figure}
\centering
\includegraphics[width=0.95\linewidth]{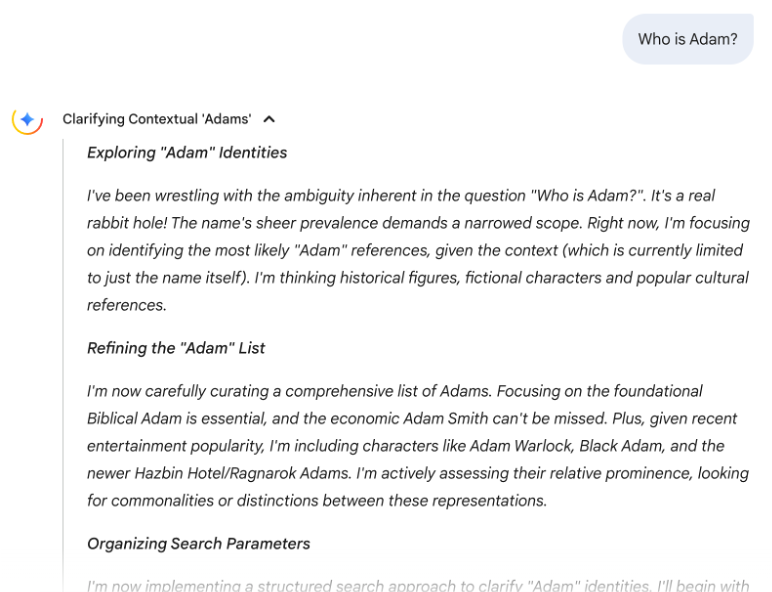}
\caption{A Simple case of LLM overthinking.}
\label{fig:simple case}
\end{figure}
\end{document}